\pdfoutput=1

\documentclass[11pt]{article}

\usepackage{acl}

\usepackage{times}  
\usepackage{latexsym}

\usepackage[T1]{fontenc}

\usepackage[utf8]{inputenc}

\usepackage{microtype}

\usepackage{inconsolata}

\usepackage{amsmath}
\usepackage{makecell}
\usepackage{multirow}
\usepackage[cjk]{kotex}
\usepackage{array}
\usepackage{booktabs}
\usepackage{graphicx}
\usepackage{array}
\usepackage{colortbl}

\usepackage[framemethod=TikZ]{mdframed}
\usepackage{xcolor}

\definecolor{cadmiumgreen}{rgb}{0.0, 0.42, 0.24}

\title{Call for Rigor in Reporting Quality of \\Instruction Tuning Data}

\author{Hyeonseok Moon, Jaehyung Seo, Heuiseok Lim$^{\dagger}$\\
  $^1$Department of Computer Science and Engineering, Korea University\\
  \texttt{\{glee889,seojae777,limhseok\}@korea.ac.kr}}

\begin{document}
\maketitle
\begin{abstract}
Instruction tuning is crucial for adapting large language models (LLMs) to align with user intentions. Numerous studies emphasize the significance of the quality of instruction tuning (IT) data, revealing a strong correlation between IT data quality and the alignment performance of LLMs. In these studies, the quality of IT data is typically assessed by evaluating the performance of LLMs trained with that data. However, we identified a prevalent issue in such practice: hyperparameters for training models are often selected arbitrarily without adequate justification. We observed significant variations in hyperparameters applied across different studies, even when training the same model with the same data. In this study, we demonstrate the potential problems arising from this practice and emphasize the need for careful consideration in verifying data quality. Through our experiments on the quality of LIMA data and a selected set of 1,000 Alpaca data points, we demonstrate that arbitrary hyperparameter decisions can make any arbitrary conclusion.

\end{abstract}

\section{Introduction}

Instruction Tuning (IT) is a widely adopted strategy for enabling a human-interactive use of the knowledge embedded in large language models (LLMs) \cite{cao2023instruction, wang2024survey}. By training with datasets composed of instruction-response pairs, LLM can attain the ability to generate appropriate responses to given instructions \cite{dubois2023alpacafarm, zheng2023judging, xu2023wizardlm, conover2023free}.

In implementing IT, data quality is considered a critical factor \cite{lima, wang2024survey, treeinst, lu2024instag}. Several studies have proven that selectively using high-quality IT data for training leads to better alignment performance than using the entire dataset \cite{liu2024what, chen2024alpagasus, zhao2024long, mekala2024smaller}.

Traditionally, the quality of IT data is measured by evaluating the performance of models trained on it \cite{liu2024what, chen2024alpagasus, zhao2024long, xia2024less}. 
This approach stems from the consensus that \textit{data is deemed good if it produces a good model}. Consequently, most studies on data quality establish a training configuration for models that represent data quality. Then, the performance of the trained model is regarded as the data quality. \cite{lima, zhao2024long, xia2024rethinking, du2023mods, zhou2023dataset}.


\begin{table*}[t]
\centering
\resizebox{0.9\linewidth}{!}{
\small
\begin{tabular}{c|cccc|c}

\toprule[1.5pt]
\textbf{Paper} & \textbf{Epochs} & \textbf{LR} & \textbf{LR Scheduler} & \textbf{Batch} & \textbf{Data Pool} \\ \midrule[1.5pt]

\multicolumn{6}{c}{\textit{Training \textbf{Llama-2-7B} with sampled \textbf{1K} general domain IT data}} \\ \midrule[1.5pt]

\citet{ghosh2024closer} & 3 & 5e-5 & - & 32 & Lima \\

\citet{raghavendra2024revisiting} & 3 & 1e-5 & - & 8 & Dolly \\

\citet{yu2024diversify} & 3 & 1e-5 & - & 64 & Alpaca / WizardLM \\

\citet{du2023mods} & 3 & 2e-5 & Cosine & 128 & Alpaca+HC3+WizardLM+Dolly+Self-Instruct+Lima\\

\citet{li-etal-2024-quantity} & 3 & 2e-5 & - & 128 & Alpaca / WizardLM \\

\citet{liu2024selectit} & 3 & 2e-5 & Cosine & 128 & Alpaca-gpt4 / Lima \\

\citet{mekala2024smaller} & 3 & 2e-5 & Cosine & 128 & Alpaca / Dolly \\

\citet{DBLP:journals/corr/abs-2407-08995} & 8 & 1e-5 & Cosine & 64 & Lima \\

\citet{zhao2024long} & 15 & 1e-5 & Linear & 128 & Alpaca / WizardLM / Lima \\ 

\citet{lima} & 15 (ES) & 1e-5 & Linear & 64 & Lima \\


 \midrule[1.5pt]

\multicolumn{6}{c}{\textit{Training \textbf{Llama-2-13B} with 1K general domain IT data }} \\ \midrule[1.5pt]

\citet{ghosh2024closer} & 3 & 5e-5 & - & 32 & Lima \\ 
 
\citet{treeinst} & 10 & 1e-4 & - & 16 & Alpaca-gpt4 \\

\citet{liu2024selectit} & 3 & 2e-5 & Cosine & 128 & Alpaca-GPT4 / Lima \\

\citet{mekala2024smaller} & 3 & 2e-5 & Cosine & 128 & Alpaca / Dolly\\

\citet{zhao2024long} & 15 & 1e-5 & Linear & 128 & Alpaca / WizardLM / Lima \\ \midrule[1.5pt]
 
\multicolumn{6}{c}{\textit{Training \textbf{Mistral 7B} with 1K general domain IT data }} \\ \midrule[1.5pt]
\citet{DBLP:journals/corr/abs-2407-08995} & 4 & 1e-5 & Cosine & 64 & Lima\\
\citet{zhao2024long} & 15 & 2e-6 & Linear & 128 & Alpaca / WizardLM / Lima \\  \citet{ghosh2024closer} & 3 & 5e-5 & - & 32 & Lima \\
\citet{yu2024diversify} & 3 & 1e-5 & - & 64 & Alpaca / WizardLM\\

\citet{entropylaw} & 4 & 4e-6 & - & 128 & WizardLM / UltraChat / ShareGPT \\
  
\bottomrule[1.5pt]

\end{tabular}
}
\caption{
Hyperparameters reported by previous studies, adopted to train LLMs with 1K general domain IT data. The data pool details the sources from which the 1K data samples were drawn. Detailed descriptions of these data pools are provided in the Table~\ref{tb:pool}. The '+' symbol indicates experiments where samples were drawn from a combined data mix of all mentioned datasets. The '/' symbol reports studies that sampled individually from each data pool.
} \label{tb:survey}
\end{table*}

However, we observed that these studies often lack justification for the hyperparameter settings used in model training. Table~\ref{tb:survey} presents the diverse hyperparameter configurations utilized in previous research implementing IT with a sampled 1K general domain IT dataset. We discovered that the configurations may vary across studies, even when training the same model with identical data sizes.

In this study, we question whether reaching coherent conclusions under varying settings is possible. Specifically, we emphasize that conclusions regarding data quality can easily be altered based on arbitrarily chosen hyperparameter settings. For instance, even if one might report that dataset A is superior to dataset B, another could claim that B is better by training models under different settings, even with the same dataset, model, and test settings. This variability poses a risk of causing significant confusion. 

As a representative case, we consider two general-domain IT datasets: LIMA \cite{lima} and sampled 1K dataset from Alpaca (Alpaca-longest \cite{zhao2024long}). In \cite{zhao2024long}, it was reported that a model trained on Alpaca-longest outperformed a model trained on LIMA. However, our experiments contrarily demonstrate that LIMA can also be regarded as better than Alpaca-longest, depending on the selected training setting. Given the current research trend of arbitrarily determining hyperparameters for validation models, this confusion can be identified as a severe yet persistent problem.

Through our experiments, we emphasize the necessity of rigor in reporting data quality. Furthermore, our discussion suggests the importance of identifying (at least) locally optimal hyperparameters and reporting data quality under these settings.

\section{Related Works}

\begin{figure*}[t]
\centering
\includegraphics[width=1.0\linewidth]{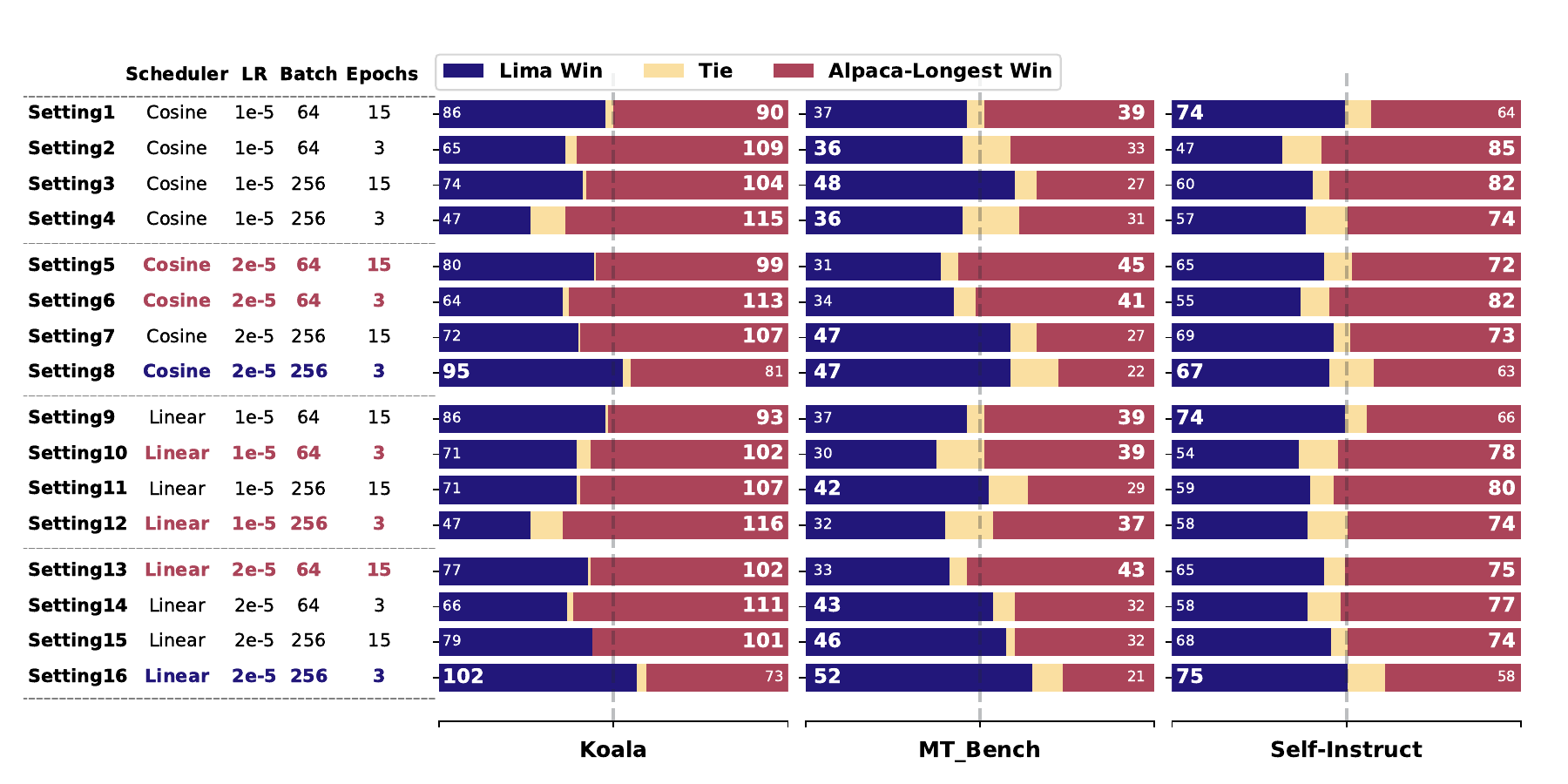}
 \caption{The performance comparison between the two models trained with LIMA and Alpaca-Longest. We train Llama-2-7B model with each dataset, We evaluate the data quality when training each dataset with the Llama-2-7B model. is depicted on the 
 Y-axis represents the hyperparameter settings used in each experiment. We bolded the settings that consistently demonstrated conclusive results across all three evaluation datasets.}
 \label{fig:lima_vs_long}
\end{figure*}

Previous research has widely acknowledged the importance of data quality in performing IT. \citet{chen2024alpagasus} proposed that training LLMs with a small, carefully selected subset of high-quality data can significantly improve alignment performance within the vast IT data pool. Furthermore, \citet{lima} even suggested that carefully curated high-quality 1,000 data points are sufficient to attain alignment performance for LLMs. Motivated by these findings, numerous studies are exploring various methodologies focused on selecting high-quality instruction tuning data \cite{wang2024survey, chen2024alpagasus, zhao2024long, xia2024rethinking, lu2024instag, liu2024what}.

However, most studies lack justification for the selected hyperparameter setting to train verification models. Consequently, the training setups become diversified even when using the same LLM and dataset. We argue that the importance of selecting appropriate hyperparameters has long been emphasized \cite{yu2020hyper, mccandlish2018empirical, halfon2024stay}. The community widely recognizes that optimal hyperparameters are often specific to particular LLMs and datasets, and reported performance may vary based on the experimental setup \cite{van2018hyperparameter, jin2022hyperparameter, gkouti-etal-2024-try, bi2024deepseek}. However, we find that research on data quality frequently reports performance without adequately considering these factors.

In this study, we highlight the potential confusion that can result from neglecting these considerations and demonstrate the necessity of a rigorous experimental setup to report data quality.



\section{Experimental Setting}

\subsection{Exam-taker Dataset}
In our experiments, we adopt two general domain IT datasets, each comprising 1,000 samples, as our exam-taker datasets. By comparing the quality of these two datasets, we examine how the judgment on the exam-taker datasets varies with different arbitrarily chosen hyperparameter settings.

\paragraph{LIMA \cite{lima}} LIMA is a high-quality dataset comprising 1,000 IT data points, carefully curated by human efforts with an emphasis on quality and diversity.

\paragraph{Alpaca-Longest \cite{zhao2024long}}
\citet{zhao2024long} selected the 1,000 entries with the longest token lengths from the Alpaca dataset \cite{alpaca}. This approach proved more effective than training on the entire Alpaca dataset and significantly outperformed other baselines such as Alpagasus \cite{chen2024alpagasus}. According to the original paper, training with this data resulted in higher alignment performance than LIMA.

\subsection{Experimental Model}

The quality of the Exam-taker Dataset is determined by the performance of the experimental model trained on it. We conduct experiments using the Llama-2-7B model \cite{touvron2023llama} and the Mistral-7B-v0.3 model \cite{jiang2023mistral}. The main paper reports the results for the Llama-2-7B model, and Appendix~\ref{app:mistral} includes the results for the Mistral-7B model.

\subsection{Experimental Setting}
This study focuses on four commonly reported hyperparameters: learning rate, learning rate scheduler, batch size, and number of epochs. We report the experimental results obtained from varying these parameters. We conduct comparative experiments for each setting by choosing two prevalent yet distinct values. While numerous other potential variations exist, such as weight decay and dropout, we leave these for future exploration. Apart from the hyperparameters under investigation, detailed experimental settings are provided in Appendix~\ref{app:experimental_detail}.

\subsection{Test Dataset}
To evaluate the performance of the trained model, we use three LLM alignment benchmarks: Koala \cite{geng2023koala}, MT-Bench \cite{zheng2023judging}, and Self-Instruct \cite{wang-etal-2023-self-instruct}. These benchmarks serve as an instruction-following evaluation tool, assessing LLMs by evaluating the quality of text generated in response to given instructions. We employ GPT-4o \cite{hurst2024gpt}\footnote{https://openai.com/index/hello-gpt-4o/} as a judge to compare the performance of experimental models for each benchmark. The judge prompts used in the experiments are detailed in Appendix~\ref{app:judge_prompt}.

\section{Experimental Results}
\label{sec:experiments}
\subsection{LIMA vs Alpaca-Longest}
Figure~\ref{fig:lima_vs_long} presents the experimental results based on hyperparameter variations. 
Our results show that if we choose specific settings (\textit{e.g.,} Settings 4, 5, 10, 12, 13), we can report that Alpaca-longest exhibits superior data quality compared to LIMA. At the same time, if we choose other configurations (\textit{e.g.,} Settings 8, 16), we can report that LIMA still demonstrates higher data quality.

Considering that authors have determined such hyperparameters arbitrarily, this represents a significant concern. We view that the ability to alter reported conclusions based on subjective decisions can severely undermine the reliability of scientific discussions.


\begin{table}[t]
\centering
\resizebox{1.0\linewidth}{!}{
\small
\begin{tabular}{l|ccc|ccc}

\toprule[1.5pt]
\makecell[c]{\textbf{Dataset}} & \multicolumn{3}{c|}{\textbf{LIMA}} & \multicolumn{3}{c}{\textbf{Alpaca-Longest}} \\ \midrule[1.5pt]

\makecell[c]{\textbf{Comparison} \\
\textbf{with} \\
\textbf{Setting 1}}
& \makecell[c]{\textbf{Setting 1} \\
\textbf{Wins}} 
& \textbf{Tie}
& \makecell[c]{\textbf{Setting} $x$ \\
\textbf{Wins}}
& \makecell[c]{\textbf{Setting 1} \\
\textbf{Wins}} 
& \textbf{Tie}
& \makecell[c]{\textbf{Setting} $x$ \\
\textbf{Wins}} \\ \midrule

\textit{vs} \textbf{Setting 2} & \textbf{112} & 10 & 58 & \textbf{121} & 7 & 52 \\
\textit{vs} \textbf{Setting 3} & \textbf{101} & 7 & 72 & \textbf{97} & 10 & 73 \\
\textit{vs} \textbf{Setting 4} & \textbf{142} & 7 & 31 & \textbf{144} & 10 & 26 \\
\textit{vs} \textbf{Setting 5} & \textbf{98} & 9 & 73 & \textbf{102} & 6 & 72 \\
\textit{vs} \textbf{Setting 6} & \textbf{114} & 6 & 60 & \textbf{109} & 7 & 64 \\
\textit{vs} \textbf{Setting 7} & 83 & 8 & \textbf{89} & 73 & 7 & \textbf{100} \\
\textit{vs} \textbf{Setting 8} & \textbf{121} & 8 & 51 & \textbf{136} & 9 & 35 \\
\textit{vs} \textbf{Setting 9} & \textbf{86} & 10 & 84 & \textbf{84} & 14 & 82 \\
\textit{vs} \textbf{Setting 10} & \textbf{116} & 9 & 55 & \textbf{130} & 9 & 41 \\
\textit{vs} \textbf{Setting 11} & \textbf{101} & 5 & 74 & \textbf{96} & 9 & 75 \\
\textit{vs} \textbf{Setting 12} & \textbf{146} & 7 & 27 & \textbf{145} & 9 & 26 \\
\textit{vs} \textbf{Setting 13} & \textbf{96} & 7 & 77 & \textbf{102} & 8 & 70 \\
\textit{vs} \textbf{Setting 14} & \textbf{124} & 8 & 48 & \textbf{114} & 5 & 61 \\
\textit{vs} \textbf{Setting 15} & 82 & 7 & \textbf{91} & 78 & 6 & \textbf{96} \\
\textit{vs} \textbf{Setting 16} & \textbf{109} & 10 & 61 & \textbf{145} & 5 & 30 \\
\bottomrule[1.5pt]

\end{tabular}
}
\caption{
We report the performance of the Llama-2-7B model, trained under each setting, as evaluated on the Koala dataset. Details for each setting are presented in the Figure~\ref{fig:lima_vs_long}.} \label{tb:setting}
\end{table}

\subsection{Among the Same Dataset}
Then, which setting should we choose to report? Considering that the primary goal of the IT dataset is to construct high-performance models, it would be reasonable and practical to report results based on the best achievable performance with the given data \cite{koehn-EtAl:2018:WMT, koehn2020findings, budach2022effects, van2018hyperparameter}. 

In this section, we identify the optimal settings among the configurations tested. We recognize that other configurations with better performance may have been overlooked. We focus on local optimality within our considered settings and discuss its implications. Figure~\ref{tb:setting} compares model performance across various hyperparameter settings, using \textbf{Setting1} as the baseline. 

Our experiments reveal that \textbf{Setting7, 15} (2e-5 LR / 256 Batch / 15 Epochs) maximizes model performance within our study. Notably, we can find that such configurations are far beyond the widely chosen settings in existing research. As our brief survey in Table~\ref{tb:survey} indicates, most studies opt to train Llama-2-7B for only three epochs when using 1K IT datasets. However, our results show that this setup yielded significantly lower performance than training for 15 epochs under the same conditions. This finding suggests that the reported performance in many studies may reflect the under-trained performance of models, which may fail to fully represent the potential of the exam-taker dataset.

\section{Discussion}

\paragraph{We argue that it is inevitable to evaluate the downstream model performance.}

We acknowledge that assessing data quality through the performance of a trained model can be ambiguous. However, we also argue that the quality of training data must inevitably be assessed through the model's performance after training. 

We would like to discuss how the quality of training data is generally acknowledged. The goal of constructing the training dataset is to develop a model that aligns with the intended purpose. Thus, in terms of training data, "good data" is defined as data that produces a "good model"\cite{chen2024alpagasus, koehn2020findings}. This fundamentally differs from constructing benchmark datasets for evaluation. Since training data's primary aim is to build a strong model, data that appear high-quality to humans (or any frontier LLMs) may offer little value if the trained model's performance remains subpar \cite{liu2024what}.

In this context, the quality of training data is fundamentally linked to the performance of the model trained on it. While there are various plausible methods to assess training data, these methods might remain indirect indicators, without validating with the performance of the trained model.

Consequently, most studies demonstrate the quality of the data under evaluation by training it onto one (possibly several) model and reporting their performance. We do not consider this approach erroneous; instead, we view it as a natural and inevitable choice. Our stance is that if model-based verification is unavoidable, a more thorough and rigorous training configuration would be essential to verify data quality. Our experiments demonstrate that hyperparameters can introduce unintended biases that skew the objective evaluation of the data quality.

\paragraph{We argue that authors researching data quality have responsibility for such validation.}


To address these ambiguities, we suggest selecting the hyperparameter setting that yields the highest performance within a given data and reporting the model's performance under this setting. This approach seeks to evaluate the model by maximizing the data's potential. 

Given that hyperparameter search is being performed in relatively small-sized PLMs (e.g., BERT \cite{devlin2019bert}, BART \cite{lewis2020bart}) \cite{latif2024evaluation, ljubesic-etal-2024-language, roele-2021-wvoq}, we argue that it is challenging to justify its omission in LLMs other than its high cost. Even when researchers do not conduct their own hyperparameter search, there have been multiple attempts to use existing configurations \cite{zhou2023dataset}. However, as shown in Table~\ref{tb:survey} with the example of Mistral, there appears to be no established standard configuration when tuning relatively recent LLMs.

Reporting the best performance would certainly require additional costs for experiments, but we believe this is a necessary sacrifice to strengthen scientific discourse. We argue that arbitrary conclusions stemming from arbitrary hyperparameter choices pose a greater risk than incurring additional costs. While a comprehensive hyperparameter search may not always be necessary, we claim that authors should clearly justify their chosen hyperparameters. Even if they do not report peak performance, employing the best settings from our paper or established training configurations (ex. LIMA configuration) would still be a rational approach.

\section{Conclusion}
In our examination of various studies addressing data quality, we observed a recurring issue where researchers often arbitrarily select hyperparameters when training models to verify data quality. Our experiments reveal that arbitrary hyperparameter choices can lead to arbitrary conclusions. Moreover, we found that hyperparameters chosen without justification often fail to achieve optimal performance on the exam-taker datasets, resulting in unreliable conclusions. To address this, we propose establishing a local hyperparameter pool and training models under locally optimal settings within this pool. While additional costs for hyperparameter validation are inevitable, we consider this a necessary sacrifice for attaining the reliability of scientific discourse. To ensure rigorous reporting and sustainable consensus, we urge careful attention.

\section*{Limitation}
The numerous hyperparameter settings we did not consider may remain a limitation of our study. Within our budget constraints, we verified as many possibilities as possible. Fortunately, we found significant variations in model performance even within the four factors we examined, allowing us to draw generalized conclusions. Through our brief survey, we can also found that various other hyperparameter variants, such as weight decay, warmup steps, are also introduced without justification. Exploring additional possibilities to identify an optimal setup could be a meaningful area for future research.

In this study, we did not attempt hyperparameter optimization (HPO), as finding optimal values was outside the scope of our research. Applying HPO when reporting on data quality, could serve as an excellent direction for future research. 

Although we conducted experiments using only two datasets, we do not see this as a limitation. We believe this setup clearly demonstrates the inherent ambiguity in reporting data quality. Possibly numerous other datasets can exist where hyperparameter settings could alter reporting conclusions. Instead of identifying additional dataset pairs, we consider it more valuable to focus future research on strategies to mitigate such ambiguities.

\section*{Ethics Statement}
We do not challenge the quality assessment results reported by the \citet{zhao2024long}, which proposed Alpaca-Longest and estimated that Alpaca-Longest is better than LIMA. Based on the results from our experiments using locally optimal settings within our pool, Alpaca-Longest can arguably be considered superior to Lima. This finding aligns with the results reported in the original paper. In all experiments, we exclusively used artifacts approved for research purposes.


\bibliography{custom}

\newpage
\appendix
\begin{figure*}[t]
\centering
\includegraphics[width=1.0\linewidth]{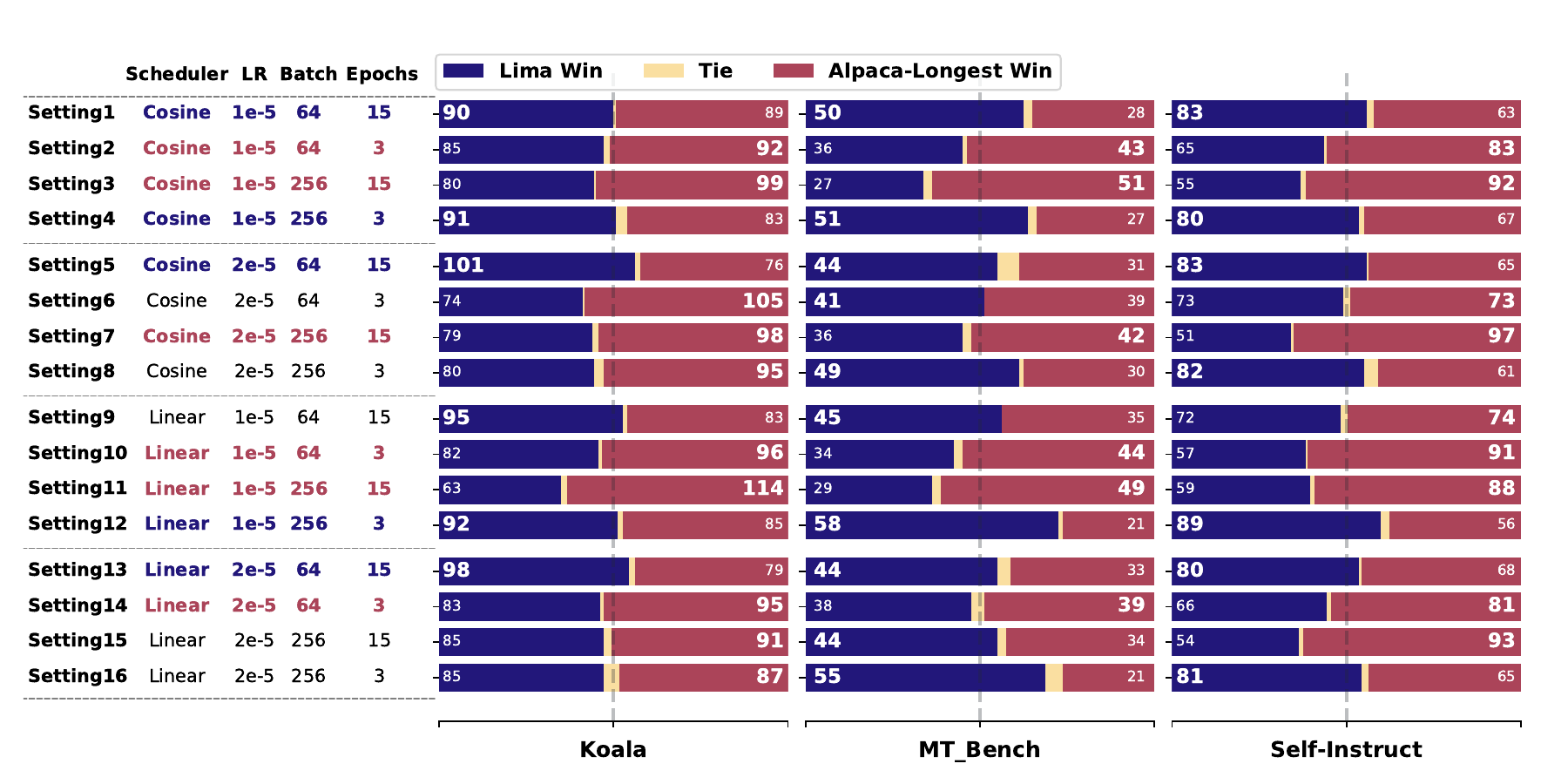}
 \caption{The performance comparison between the two models trained with LIMA and Alpaca-Longest. We train Mistral-7B model with each dataset, We evaluate the data quality when training each dataset with the Mistral-7B model. is depicted on the 
 Y-axis represents the hyperparameter settings used in each experiment. We bolded the settings that consistently demonstrated conclusive results across all three evaluation datasets.}
 \label{fig:lima_vs_long_mistral}
\end{figure*}

\section{Experimental Results - Mistral}
\label{app:mistral}
We conducted the same experiments described in Section~\ref{sec:experiments} using the Mistral-7B model. The results are reported in Table~\ref{tb:setting_mistral} and Figure~\ref{fig:lima_vs_long_mistral}.


\begin{table}[h]
\centering
\resizebox{1.0\linewidth}{!}{
\small
\begin{tabular}{l|ccc|ccc}

\toprule[1.5pt]
\makecell[c]{\textbf{Dataset}} & \multicolumn{3}{c|}{\textbf{LIMA}} & \multicolumn{3}{c}{\textbf{Alpaca-Longest}} \\ \midrule[1.5pt]

\makecell[c]{\textbf{Comparison} \\
\textbf{with} \\
\textbf{Setting 1}}
& \makecell[c]{\textbf{Setting 1} \\
\textbf{Wins}} 
& \textbf{Tie}
& \makecell[c]{\textbf{Setting} $x$ \\
\textbf{Wins}}
& \makecell[c]{\textbf{Setting 1} \\
\textbf{Wins}} 
& \textbf{Tie}
& \makecell[c]{\textbf{Setting} $x$ \\
\textbf{Wins}} \\ \midrule

\textit{vs} \textbf{Setting 2} & 79 & 12 & \textbf{89} & 58 & 11 & \textbf{111} \\
\textit{vs} \textbf{Setting 3} & 75 & 12 & \textbf{93} & 50 & 14 & \textbf{116} \\
\textit{vs} \textbf{Setting 4} & \textbf{100} & 8 & 72 & \textbf{107} & 14 & 59 \\
\textit{vs} \textbf{Setting 5} & \textbf{116} & 13 & 51 & \textbf{126} & 13 & 41 \\
\textit{vs} \textbf{Setting 6} & \textbf{97} & 17 & 66 & \textbf{95} & 15 & 70 \\
\textit{vs} \textbf{Setting 7} & \textbf{93} & 12 & 75 & 70 & 14 & \textbf{96} \\
\textit{vs} \textbf{Setting 8} & \textbf{125} & 8 & 47 & \textbf{130} & 12 & 38 \\
\textit{vs} \textbf{Setting 9} & \textbf{87} & 16 & 77 & \textbf{82} & 18 & 80 \\
\textit{vs} \textbf{Setting 10} & \textbf{83} & 14 & \textbf{83} & 58 & 15 & \textbf{107} \\
\textit{vs} \textbf{Setting 11} & 81 & 7 & \textbf{92} & 45 & 6 & \textbf{129} \\
\textit{vs} \textbf{Setting 12} & \textbf{102} & 10 & 68 & \textbf{96} & 9 & 75 \\
\textit{vs} \textbf{Setting 13} & \textbf{124} & 13 & 43 & \textbf{129} & 18 & 33 \\
\textit{vs} \textbf{Setting 14} & \textbf{96} & 22 & 62 & \textbf{87} & 16 & 77 \\
\textit{vs} \textbf{Setting 15} & \textbf{88} & 14 & 78 & 78 & 14 & \textbf{88} \\
\textit{vs} \textbf{Setting 16} & \textbf{109} & 9 & 62 & \textbf{118} & 10 & 52 \\

\bottomrule[1.5pt]

\end{tabular}
}
\caption{
We report the performance of the Mistral-7B model, trained under each setting, as evaluated on the Koala dataset. Details for each setting are presented in the Figure~\ref{fig:lima_vs_long}.} \label{tb:setting_mistral}
\end{table}

As shown in Table~\ref{tb:setting_mistral}, models trained for 15 epochs generally outperformed those trained for only 3 epochs, even within the same settings. This finding suggests that commonly adopted hyperparameter settings in prior research may not be optimal and that reported performance might not fully exploit the data's potential.

Figure~\ref{fig:lima_vs_long_mistral} illustrates the potential conclusions we can draw from various settings using Mistral. There is still significant diversity between settings, supporting our earlier conclusions in Section~\ref{sec:experiments}. We demonstrate that merely using multiple models is insufficient to enhance robustness in data quality validation, emphasizing the necessity of hyperparameter generalization.

\section{Dataset Details}
\label{app:dataset_detail}


\begin{table}[h]
\centering
\resizebox{1.0\linewidth}{!}{
\small
\begin{tabular}{c|c|c}

\toprule[1.5pt]
\textbf{Dataset} & \textbf{Paper / Description} & \textbf{Data Size} \\ \midrule[1.5pt]
Alpaca & \citet{alpaca} & 52K \\
Alpaca-GPT4 & \citet{peng2023instruction} & 52K \\
Dolly & \citet{conover2023free} & 15K \\
HC3 & \citet{guo2023close} & 24.3K \\
ShareGPT & \citet{chiang2023vicuna} & 52K \\
UltraChat & \citet{ding-etal-2023-enhancing} & 200K \\
WizardLM & \citet{xu2023wizardlm} & 700K \\
\bottomrule[1.5pt]

\end{tabular}
}
\caption{
We report only the data aimed at performing IT in a general domain, which are adopted to previous studies. Each dataset consists of a pair, featuring a human instruction and an appropriate response.} \label{tb:pool}
\end{table}

\section{Experimental Details}
\label{app:experimental_detail}
We conducted experiments with a weight decay of 0.0, a warmup of 0.0, and a maximum length of 2,048, utilizing the \texttt{HuggingFace} trainer \cite{wolf-etal-2020-transformers}. To enhance learning efficiency, we applied \texttt{bf16} \cite{kalamkar2019study} and \texttt{tf32} \cite{stosic2021accelerating} strategy. All training was performed using FlashAttention-2 \cite{flashattn2} and DeepSpeed Stage 2 \cite{smith2022using}. For inference, we employed vllm \cite{kwon2023efficient}. Our setup included four RTX-A6000 GPUs with 48GB each for model training and inference. The original batch size per GPU was set to 2, and we used gradient accumulation to increase the batch size. Other settings followed the default configurations provided by the \texttt{HuggingFace} trainer. 

\section{LLM-as-a-Judge}
\label{app:judge_prompt}

\definecolor{judgecolor}{HTML}{F5F5F7}

\begin{table}[h]
\centering
\resizebox{1.0\linewidth}{!}{
\begin{tabular}{l}

\toprule[1.5pt]
\rowcolor{judgecolor} \makecell[l]{
\textbf{\#\# System Prompt} \\
Please act as an impartial judge and evaluate the quality \\
of the responses provided by two AI assistants to the \\
user question displayed below. \\
You should choose the assistant that follows the user’s \\
instructions and answers the user’s question better. \\
Your evaluation should consider factors such as the \\
helpfulness, relevance, accuracy, depth, creativity, \\
and level of detail of their responses. \\
Begin your evaluation by comparing the two responses \\ 
and provide a short explanation. Avoid any position \\
biases and ensure that the order in which the responses \\
were presented does not influence your decision. \\
Do not allow the length of the responses to influence \\
your evaluation. Do not favor certain names of the \\
assistants. Be as objective as possible. \\
After providing your explanation, output your final \\
verdict by strictly following this format: "[[A]]" \\
if assistant A is better, "[[B]]" if assistant B is \\
better, and "[[C]]" for a tie. \\
\textbf{\#\# Input Statements} \\
You are a helpful and precise assistant for \\
checking the quality of the answer. \\
\text{[Question]} \\
\textit{\{question\}} \\
\text{[The Start of Assistant 1's Answer]} \\
\textit{\{Response From Assistant 1\}} \\
\text{[The End of Assistant 1's Answer]} \\
\text{[The Start of Assistant 2's Answer]} \\
\textit{\{Response From Assistant 2\}} \\
\text{[The End of Assistant 2's Answer]} \\
} \\
\bottomrule[1.5pt]

\end{tabular}} \caption{Prompt used for training the LLM: For models not supporting system prompts, we combined the system prompt and user prompt into a single input statement.} \label{tb:prompt_judge}
\end{table}

The prompt we used is presented in Table~\ref{tb:prompt_judge}. In all our experiments, we randomize the order of presented responses to relieve any unintended effects driven by the positional bias. We conducted our experiments with GPT-4o (\texttt{gpt-4o-2024-08-06}), setting the temperature to 0 and top-p to 1.0. The API usage cost for the experiments detailed in Table~\ref{tb:setting} was \$11.25. Conducting a similar hyperparameter search using GPT4o-mini incurred a cost of \$1.44. While using GPT4o-mini presents a cost-effective option, the Pearson-r correlation score between GPT4o and GPT4o-mini was 0.559 in our experiments. Although this score might be considered reasonably high, we argue using GPT-4o is more effective for establishing a more precise and rigorous setting. We leave experiments with alternative judges for future research.

\end{document}